\documentclass{article}

\usepackage{microtype}
\usepackage{graphicx}
\usepackage{subcaption}
\usepackage{booktabs} 

\usepackage{hyperref}

\usepackage[preprint]{icml2026}

\usepackage{amsmath}
\usepackage{amssymb}
\usepackage{mathtools}
\usepackage{amsthm}

\usepackage[capitalize,noabbrev]{cleveref}

\theoremstyle{plain}
\newtheorem{theorem}{Theorem}[section]

\theoremstyle{definition}
\newtheorem{definition}[theorem]{Definition}

\theoremstyle{remark}

\usepackage[textsize=tiny]{todonotes}

\icmltitlerunning{High-Dim Search, Low-Dim Solution}

\begin{document}

\twocolumn[
  \icmltitle{High-Dimensional Search, Low-Dimensional Solution: Decoupling Optimization from Representation
}



  \icmlsetsymbol{equal}{*}

  \begin{icmlauthorlist}
    \icmlauthor{Yusuf Kalyoncuoglu}{equal,yyy}
    \icmlauthor{Ratmir Miftachov}{equal,xxx}
  \end{icmlauthorlist}

  \icmlaffiliation{yyy}{Department of Computer Science, RWTH Aachen University, Aachen, Germany}
  \icmlaffiliation{xxx}{Institute of Mathematics, School of Business and Economics, Humboldt-Universität zu Berlin, Berlin, Germany}

  \icmlcorrespondingauthor{Ratmir Miftachov}{ratmir.miftachov@hu-berlin.de}

  \icmlkeywords{Machine Learning, ICML}

  \vskip 0.3in
]



\printAffiliationsAndNotice{\icmlEqualContribution}


\begin{abstract}
State-of-the-art models rely on massive widths despite exhibiting low Intrinsic Dimension (ID). We posit that this redundancy serves the non-convex \textit{optimization search} rather than the final representation. We validate this hypothesis by decoupling the solution geometry via data-independent random projections, demonstrating that ResNet, ViT, and BERT representations can be compressed by up to $16\times$ with negligible performance degradation of around 1\%. Notably, these oblivious projections achieve parity with PCA and learned baselines, confirming the solution manifold is intrinsically robust.
These findings establish the foundation for \textit{Subspace-Native Distillation}: a paradigm where student models target this intrinsic manifold directly, bypassing the high-dimensional optimization bottleneck to realize the vision of ``Train Big, Deploy Small''.
\end{abstract}

\section{Introduction}
\label{sec:intro}

The scaling laws of deep learning dictate that larger models with higher-dimensional widths generally yield better performance \citep{kaplan2020scaling}. Consequently, state-of-the-art architectures such as Transformers in NLP and Vision operate with embedding dimensions ranging from $d = 768$ (Vision-Transformer, \citet{dosovitskiy2020vit}) to over $d = 16384$ (Llama 3.1, \citet{grattafiori2024llama}). While this over-parameterization aids the optimization process by smoothing the loss landscape \citep{li2018visualizing}, it raises a central question regarding inference efficiency and representation geometry: Does achieving state-of-the-art accuracy require \textit{high-dimensional representations}, or is this dimensionality merely a temporary aid for the \textit{optimization process}?

Empirical evidence points to the latter, suggesting that the redundancy is structural, rooted in the spectral properties of the weight matrices themselves. Analyzing the Empirical Spectral Density (ESD) of deep networks, \citet{martin2021implicit} identified a phenomenon of ``Heavy-Tailed Self-Regularization.'' The authors demonstrate that well-trained weight matrices do not utilize their full algebraic rank. Instead, the information concentrates along a small number of dominant singular vectors, while the vast majority of dimensions constitute noise or bulk components, which are indistinguishable from random noise under the Marchenko-Pastur distribution \citep{martin2021implicit}.

This spectral sparsity implies that the network scales the signal only along very few critical directions. This observation holds across architectures, including Convolutional Neural Networks, where the spectral properties have been extensively studied \citep{sedghi2018singular}. Specifically, it has been shown that the singular values of convolution operators decay sharply, allowing for aggressive low-rank approximations without retraining \citep{idelbayev2020low}.

It is precisely this low-rank geometry that enables the success of modern compression and adaptation techniques. Methods like Singular Value Decomposition (SVD) \citep{denton2014exploiting} and Low-Rank Adaptation (LoRA) \citep{hu2021lora} work effectively because they exploit the fact that weight matrices and updates reside in low-dimensional subspaces. Essentially, these methods acknowledge that the model scales along very few critical directions.

This leads to a paradox: If the final solution lives in a low-dimensional subspace, why do we not simply design and train smaller backbones from scratch? The \textit{Lottery Ticket Hypothesis} \citep{frankle2018lottery} provides a critical insight into this: it posits that large, dense networks contain sparse subnetworks, which are winning tickets, that are capable of matching the full model's accuracy when trained in isolation. The high-dimensional parameter space effectively serves as a vast combinatorial pool, significantly increasing the probability that at least one such optimal subnetwork exists at initialization and can be successfully uncovered by gradient descent.

This is further formalized by \textit{Neural Tangent Kernel} (NTK) theory \citep{jacot2018neural}, which shows that in the infinite-width limit, deep networks transition into a lazy training regime. In this state, the network's evolution is governed by a static kernel, effectively linearizing the optimization landscape and guaranteeing convergence to a global minimum \citep{lee2019wide}. Small networks, however, operate far from this regime; their optimization landscapes remain highly non-convex and chaotic, making it significantly harder to find the global minimum despite having the theoretical \textit{representational capacity} to hold the solution. Essentially, high dimensionality provides the optimization capacity required to make the search tractable.

Knowledge Distillation (KD) attempts to bridge this gap. In this framework, a Teacher refers to a large, high-dimensional pre-trained model, while a Student denotes a compact, lower-dimensional network. KD uses the Teacher to guide the Student's optimization trajectory \citep{hinton2015distilling, phuong2019towards}. However, standard KD methods typically force the student to mimic the teacher's full high-dimensional output logits or feature maps \citep{romero2014fitnets}. This imposes an unnecessarily hard constraint: the student must approximate the entire ambient space of the teacher—including its noise and redundant bulk components. This capacity gap can hinder effective transfer if the student struggles to model the teacher's high-dimensional complexity \citep{cho2019efficacy}.

Consequently, the high dimensionality of SOTA models serves primarily to facilitate this search. We propose a direct validation of this premise by asking: Can we directly construct the solution in a much smaller space? By "constructing the solution subspace", we refer to the explicit transformation of the backbone's final latent activations into a fixed, lower-dimensional space via a data-independent projection. 
This reduced space serves as the definitive space where the classification task is solved, proving that the high-dimensional ambient space is not required for linear separability.

Proving the existence of such a robust, constructible subspace would imply that the complexity lies in the \textit{process}, not the \textit{result}. By isolating this target geometry, we effectively transform the daunting search problem into a clearer regression task. If the exact low-dimensional target subspace is known, a student backbone can be optimized to construct this specific subspace directly.

To validate this hypothesis, we propose a constructive diagnostic approach. Instead of searching for an optimal subspace via data-dependent methods, we utilize fixed, data-independent \textit{Johnson-Lindenstrauss (JL)} projections \citep{johnson1984extensions}. This allows us to rigorously test whether the solution manifold is linearly separable in a random low-dimensional basis, independent of the backbone's optimization trajectory.

While our motivation stems from the massive scaling of modern LLMs, we conduct our empirical analysis on foundational, well-studied architectures (ResNet, BERT, ViT). These models serve as standard testbeds in the theoretical literature \citep{papyan2020prevalence, martin2021implicit}, allowing us to isolate the phenomenon of geometric compression without the confounding variables of quantization or mixture-of-experts layers often present in billion-scale models.

Our contributions are as follows:
\begin{itemize}
\item We perform a systematic evaluation of fixed subspace classification across three distinct architectures and modalities: BERT \citep{devlin2018bert} on MNLI \citep{williams2018broad}, ViT \citep{dosovitskiy2020vit} on ImageNet-100 \citep{tian2020contrastive}, and ResNet-50 \citep{he2016deep} on CIFAR-100 \citep{krizhevsky2009learning}.
\item We systematically map the performance of the solution space across multiple reduction factors. Our results show that even under extreme compression of up to $16\times$ (e.g., $2048 \to 128$ or $768 \to 64$), the model exhibits minimal accuracy loss ($\approx 1\%$). This consistent stability across various dimensions proves that the intrinsic solution is not only low-dimensional but also remarkably robust to random geometric contraction.
\item We demonstrate that our oblivious random projections perform on par with data-dependent PCA and optimization-based learned projections. It shows that the discriminative signal is structurally dominant over the noise, making precise eigen-alignment redundant.

\item We establish the existence of a valid solution subspace. Based on this, we propose \textit{Subspace-Native Distillation} as a future paradigm, where student models are trained to directly target this low-dimensional manifold, bypassing the redundancy of high-dimensional teacher representations.
\end{itemize}

\section{Related Work}
\label{sec:related_work}

Our work sits at the intersection of geometric analysis of deep representations, the theory of neural collapse, and dimensionality reduction via random projections.

\subsection{Intrinsic Dimensionality of Deep Representations}
Despite the massive parameter count of modern networks, growing evidence suggests that the effective operational dimension is surprisingly low. \citet{ansuini2019intrinsic} analyzed the intrinsic dimension of intermediate layers, showing that while the ID is high in early layers, it drastically collapses in the final layers, forming a curved to flat transition. Similarly, \citet{pope2021intrinsic} demonstrated that the intrinsic dimension of image manifolds sets a fundamental limit on the generalization capability of the model.
While these works focus on \textit{measuring} the dimension using estimators (e.g., nearest-neighbor distances), our work focuses on \textit{constructing} a tangible subspace that suffices for classification.

\subsection{Neural Collapse and Spectral Properties}
The geometric structure of the final classification layer has been formalized under the framework of ``Neural Collapse'' \citep{papyan2020prevalence}. They observed that as training progresses towards zero training error, the within-class variability of features vanishes, and class means converge to the vertices of a low rank structure. This implies that the optimal separating hyperplanes lie in a low-rank subspace determined solely by the number of classes, rather than the ambient dimension.
Complementing this, \citep{martin2021implicit} applied Random Matrix Theory to show that well-trained weight matrices exhibit heavy-tailed spectral densities, where information is concentrated in a few dominant singular values. Our use of random projections empirically validates these theories: the fact that the JL projections preserve separability confirms that the signal-to-noise ratio is dominated by these few heavy-tailed components.

\subsection{Random Projections and Model Compression}
The Johnson-Lindenstrauss Lemma (JLL) \citep{johnson1984extensions} has historically been used in machine learning for speeding up kernel machines \citep{rahimi2007random} or for compressed sensing \citep{donoho2006compressed}. In the context of Deep Learning, \citet{li2018measuring} utilized random projections to measure the intrinsic dimension of the objective landscape, showing that optimization can occur in a randomly oriented low-dimensional subspace.
However, most compression techniques, such as PCA-based reduction \citet{pearson1901lines} or autoencoders \citep{hinton2006reducing}, rely on \textit{data-dependent} bases. In contrast, our approach utilizes \textit{oblivious} (data-independent) projections. By empirically proving that a fixed random basis performs on par with the full model, we show that the representation is robust enough to survive aggressive geometric distortion.

\section{Methodology}
\label{sec:method}

\subsection{Motivation: Manifold Collapse}
Let $\mathcal{X}$ denote the high-dimensional input space (e.g., image pixels or token sequences), and let $f_\theta: \mathcal{X} \to \mathbb{R}^d$ denote the pre-trained Teacher backbone (e.g., BERT or ResNet) with parameters $\theta$. While the ambient output dimension $d$ is large (e.g., 768 or 2048), \citet{ansuini2019intrinsic} demonstrate that the local intrinsic dimension of the data manifold induced by $f_\theta(\mathcal{X})$ drops significantly in the final layers. Furthermore, the Neural Collapse phenomenon implies that feature representations contract to their respective class means, forming a geometric structure that is maximally separable and intrinsically low-dimensional \citep{papyan2020prevalence}.

These findings imply that the high-dimensional embedding produced by the backbone is sparse in information content. The effective solution resides in a lower-dimensional subspace, defined in the following. 

\begin{definition}[Effective Solution Subspace]
\label{def:solution_subspace}
Let the backbone $f_\theta$ yield the high-dimensional feature representations $h \in \mathbb{R}^d$. We define the Effective Solution Subspace $\mathcal{S} \subset \mathbb{R}^d$ as a linear subspace of dimension $k \ll d$ that captures the intrinsic geometry of the data manifold.

For a tolerance $\varepsilon \ge 0$, $\mathcal{S}$ is considered a valid solution subspace if the minimum loss achievable by a linear classifier $g$ restricted to $\mathcal{S}$ is within an $\varepsilon$-margin of the minimum loss achievable in the ambient space:
\begin{equation}
    \min_{g} \mathcal{L}(g(P_{\mathcal{S}} h), y) \le \min_{f} \mathcal{L}(q(h), y) + \varepsilon
\end{equation}
where $P_{\mathcal{S}}$ is the linear projection onto $\mathcal{S}$, and $q$ is a linear classifier on the full space $\mathbb{R}^d$.
\end{definition}

Our goal is to isolate $\mathcal{S}$ directly. If the manifold is indeed flattened and low-rank, a random linear projection should be sufficient to capture its geometry without the need for learned, non-linear autoencoders.

\subsection{Constructing the Solution via Johnson-Lindenstrauss}
To construct this solution subspace, we employ random projections rooted in the Johnson-Lindenstrauss Lemma (JLL) \citep{johnson1984extensions}. We choose specifically because it provides a theoretical guarantee for preserving the Euclidean geometry of a low-dimensional manifold embedded in a high-dimensional space, irrespective of the specific basis vectors.

The JLL states that for a set of $N$ points in high-dimensional space, a random projection into $k = O(\varepsilon^{-2} \log N)$ dimensions preserves pairwise distances within a factor of $(1 \pm \varepsilon)$. Furthermore, \citet{arriaga2006algorithmic} extended this to learning theory, proving that if the data is linearly separable with a sufficient margin in the ambient space, this separability is preserved in the projected subspace with high probability.

\begin{definition}[Projection Operator]
\label{def:projection_operator}
We define the projection operator $\Phi: \mathbb{R}^d \to \mathbb{R}^k$ via a random matrix $R \in \mathbb{R}^{k \times d}$. Each entry is drawn independently from a standard normal distribution for $1 \leq m \leq k, \, 1 \leq n \leq d$:
\begin{equation}
    R_{mn} \sim \mathcal{N}(0, 1)
\end{equation}
The low-dimensional solution vector $\tilde{h} \in \mathbb{R}^k$ is obtained by projecting the Teacher's output $h$:
\begin{equation}
    \tilde{h} = \Phi(h) = \frac{1}{\sqrt{k}} R h
\end{equation}
where the scaling factor $\frac{1}{\sqrt{k}}$ ensures the preservation of the expected Euclidean norm.
\end{definition}

Crucially, $R$ is generated at initialization and remains frozen. This ensures that we are strictly assessing the intrinsic separability of the Teacher's representation. By using an oblivious (data-independent) projection, we prove that the separability is a property of the data geometry itself, not an artifact of a learned compression layer.

\subsection{Subspace Classification as Target Construction}
We treat the projected vector $\tilde{h}$ as the representation within our effective solution subspace. To validate that this subspace contains all necessary semantic information, we train a lightweight linear classifier $g_W: \mathbb{R}^k \to \mathbb{R}^C$ directly on $\tilde{h}$.

The training objective minimizes the Cross-Entropy (CE) loss over the dataset $\mathcal{D} = \{(x_i, y_i)\}_{i=1}^N$, with dataset size $N$ and number of classes $C$:
\begin{equation}
    \min_{W, b} \sum_{(x_i, y_i) \in \mathcal{D}} \mathcal{L}_{CE}\left( \text{softmax}(W \tilde{h}_i + b), y_i \right)
\end{equation}
where $W\in \mathbb{R}^{C \times k}$, $b \in \mathbb{R}^{C}$ are parameters to be learned and $\tilde{h}_i = \Phi(f_\theta(x_i))$ is the projected feature vector for the $i$-th sample.

During this phase, the Teacher backbone parameter $\theta$ are strictly frozen. Thus, $g_W$ acts as a proxy for the minimal complexity required to solve the task. If $g_W$ achieves performance parity with the full-dimensional model, it confirms that we have successfully constructed a valid solution subspace. This $\tilde{h}$ can subsequently be viewed as the intrinsic "ground truth" target for future student models, decoupling the solution from the high-dimensional optimization path of the teacher.

\section{Experimental Setting}

\subsection{Implementation Details}
In the following, we describe the models and the preprocessing/optimization configurations. To ensure reproducibility, our complete source code is provided in the supplementary material. All pre-trained model checkpoints and datasets were sourced via the HuggingFace Hub.

We implement the proposed framework across three distinct modalities using the following protocol: First, we do full fine-tuning, where all model parameters (backbone and head) are updated end-to-end. This establishes the performance ceiling of the architecture. Next, we freeze the fine-tuned backbone weights and train a linear classifier on the full-dimensional embeddings with dimension $d$. This results in the \textbf{Fine-tuned Baseline} without any geometric compression. 
Our method, the \textbf{JL Projection}, similarly builds on the full fine-tuning, but projects the embedding into the low-dimensional target with dimension $k$ via our random operator and trains a linear classifier on it. Thus, we ensure a direct comparison between the Fine-tuned Baseline and the JL Projection.

All experiments were conducted with a fixed random seed ($s=42$) to ensure deterministic behavior of the JLL projection matrices. The hyperparameters are summarized in Table \ref{tab:hyperparams}.

\begin{table}[htpb]
\centering
\caption{\textbf{Hyperparameter Configuration.} Settings for the linear probe (frozen backbone). These hyperparameters apply to both the full-dimensional baseline ($d$) and the low-dimensional subspace projections ($k$), ensuring a fair comparison. Note the use of higher learning rates (LR) for Transformers to adapt the linear head to the fixed manifold.}
\label{tab:hyperparams}
\footnotesize 
\setlength{\tabcolsep}{2.5pt} 
\begin{tabular}{@{}lccccc@{}}
\toprule
\textbf{Model} & \textbf{Optim.} & \textbf{LR} & \textbf{Wt. Decay} & \textbf{Epochs} & \textbf{Batch} \\ \midrule
ResNet-50 & SGD & $1 \times 10^{-2}$ & $5 \times 10^{-4}$ & 5 & 128 \\
BERT-base & AdamW & $1 \times 10^{-3}$ & $1 \times 10^{-2}$ & 3 & 32 \\
ViT-B/16 & AdamW & $1 \times 10^{-3}$ & $1 \times 10^{-4}$ & 3 & 32 \\ \bottomrule
\end{tabular}
\end{table}

\textbf{1. Vision (CNN): ResNet-50 on CIFAR-100}
We utilize a ResNet-50 backbone and fine-tune it on CIFAR-100. The feature vector is extracted from the final pooling layer ($d=2048$) before the fully connected layer.
\begin{itemize}
    \item \textit{Preprocessing:} We employ data augmentation to prevent overfitting on the small $32 \times 32$ CIFAR images, including random crop, horizontal flip, color jitter, and cutout regularization \citep{devries2017improved} (1 hole, max size 8).
    \item \textit{Optimization:} The model is optimized using stochastic gradient descent (SGD) with momentum (0.9). For the Subspace phase, we train for 5 epochs with a learning rate of $1e^{-2}$ and weight decay of $5e^{-4}$.
\end{itemize}

\textbf{2. NLP (Transformer): BERT-base on MNLI}
We employ the \texttt{bert-base-uncased} model ($d=768$) and fine-tune it on MNLI. The input consists of premise-hypothesis pairs with a maximum sequence length of 128 tokens.
\begin{itemize}
    \item \textit{Feature Extraction:} We utilize the \texttt{pooler\_output} (the embedding of the [CLS] token processed by a dense layer and Tanh activation) as the input $h$ for our projections.
    \item \textit{Optimization:} We use the AdamW optimizer \citep{loshchilov2017decoupled}. While full fine-tuning requires a conservative learning rate ($2e^{-5}$), the subspace classification phase allows for a higher learning rate ($1e^{-3}$) to rapidly converge the linear head on the frozen features.
\end{itemize}

\textbf{3. Vision (Transformer): ViT-B/16 on ImageNet-100}
We utilize a Vision Transformer (ViT-B/16) and fine-tune it on ImageNet-100.
\begin{itemize}
    \item \textit{Preprocessing:} Images are resized to $224 \times 224$. Similar to ResNet, we apply cutout (max. size 32) during training to encourage robust feature learning. The [CLS] token of the last hidden state serves as the feature vector.
    \item \textit{Optimization:} Training is performed with AdamW. As with BERT, we utilize a learning rate of $1e^{-3}$ for the subspace classifiers to effectively regress the target decision boundaries within 3 epochs.
\end{itemize}

\textbf{Dimensionality Reduction Factors.} We vary the target dimension $k$ across a wide spectrum. 
For ResNet-50 ($d=2048$), we evaluate subspaces of width $k \in \{1024, 512, 256, 128\}$, corresponding to compression factors up to $16\times$. 
For BERT and ViT ($d=768$), we evaluate $k \in \{512, 256, 128, 64\}$, reaching compression factors of $12\times$.

\section{Empirical Evaluation}
\label{sec:experiments}

In this section, we present the experimental results of constructing solution subspaces via Johnson-Lindenstrauss projections. We evaluate our hypothesis across three distinct modalities and architectures to ensure the universality of the phenomenon.

\subsection{Main Results}

Table \ref{tab:main_results} summarizes the classification accuracy across all three tasks. We report the raw accuracy and the relative difference compared to the full-dimensional fine-tuned baseline.

\begin{table}[htpb]
\centering
\caption{\textbf{Subspace Classification Results.} Comparison of Fine-tuned Baseline (full dimension), and JL subspace projections. $\Delta$ denotes the performance difference relative to the Baseline. Note the stable accuracy even at high compression rates (e.g., $12\times$).}
\label{tab:main_results}
\setlength{\tabcolsep}{3.5pt} 
\begin{tabular}{@{}llccrc@{}}
\toprule
Method & Dim ($k$) & Ratio & Acc. & $\Delta$ Base \\ 
\midrule
\multicolumn{5}{@{}l}{\textit{ResNet-50 (CIFAR-100)}} \\ 
\midrule
\textbf{Fine-tuned Baseline} & \textbf{2048} & \textbf{$1\times$} & \textbf{82.40\%} & \textbf{--} \\
JL Projection & 1024 & $2\times$ & 81.32\% & -1.08\% \\
JL Projection & 512 & $4\times$ & 81.20\% & -1.20\% \\
JL Projection & 256 & $8\times$ & 80.89\% & -1.51\% \\
JL Projection & 128 & $16\times$ & 80.19\% & -2.21\% \\ 
\midrule
\multicolumn{5}{@{}l}{\textit{ViT-B/16 (ImageNet-100)}} \\ 
\midrule
\textbf{Fine-tuned Baseline} & \textbf{768} & \textbf{$1\times$} & \textbf{94.02\%} & \textbf{--} \\
JL Projection & 512 & $1.5\times$ & 93.52\% & -0.50\% \\
JL Projection & 256 & $3\times$ & 93.98\% & -0.04\% \\
JL Projection & 128 & $6\times$ & 93.90\% & -0.12\% \\
JL Projection & 64 & $12\times$ & 93.28\% & -0.74\% \\ 
\midrule
\multicolumn{5}{@{}l}{\textit{BERT-base (MNLI)}} \\ 
\midrule
\textbf{Fine-tuned Baseline} & \textbf{768} & \textbf{$1\times$} & \textbf{83.74\%} & \textbf{--} \\
\textbf{JL Projection} & \textbf{512} & \textbf{$1.5\times$} & \textbf{83.93}\% & \textbf{+0.19}\% \\
JL Projection & 256 & $3\times$ & 83.79\% & +0.05\% \\
JL Projection & 128 & $6\times$ & 83.79\% & +0.05\% \\
JL Projection & 64 & $12\times$ & 83.64\% & -0.10\% \\ 
\bottomrule
\end{tabular}
\end{table}

\textbf{Robustness of the Solution Subspace:}
Across all architectures, the performance drop is minimal even under aggressive compression. For BERT (and ViT), reducing the dimension from 768 to 64 (a $12\times$ reduction) results in less than $1\%$ accuracy drop. Similarly, ResNet-50 retains over $97\%$ of its baseline performance when projected down to 128 dimensions. This confirms our hypothesis that the intrinsic solution manifold is significantly smaller than the ambient space allocated by the architecture. In fact, for BERT, mild compression ($k=512$) yields a slight accuracy gain ($+0.19\%$), implying that the ambient representation contains noise components that the projection filters out.

\begin{figure}[H]
    \centering
    \begin{subfigure}{0.49\linewidth}
        \centering
        \includegraphics[width=1.0\linewidth]{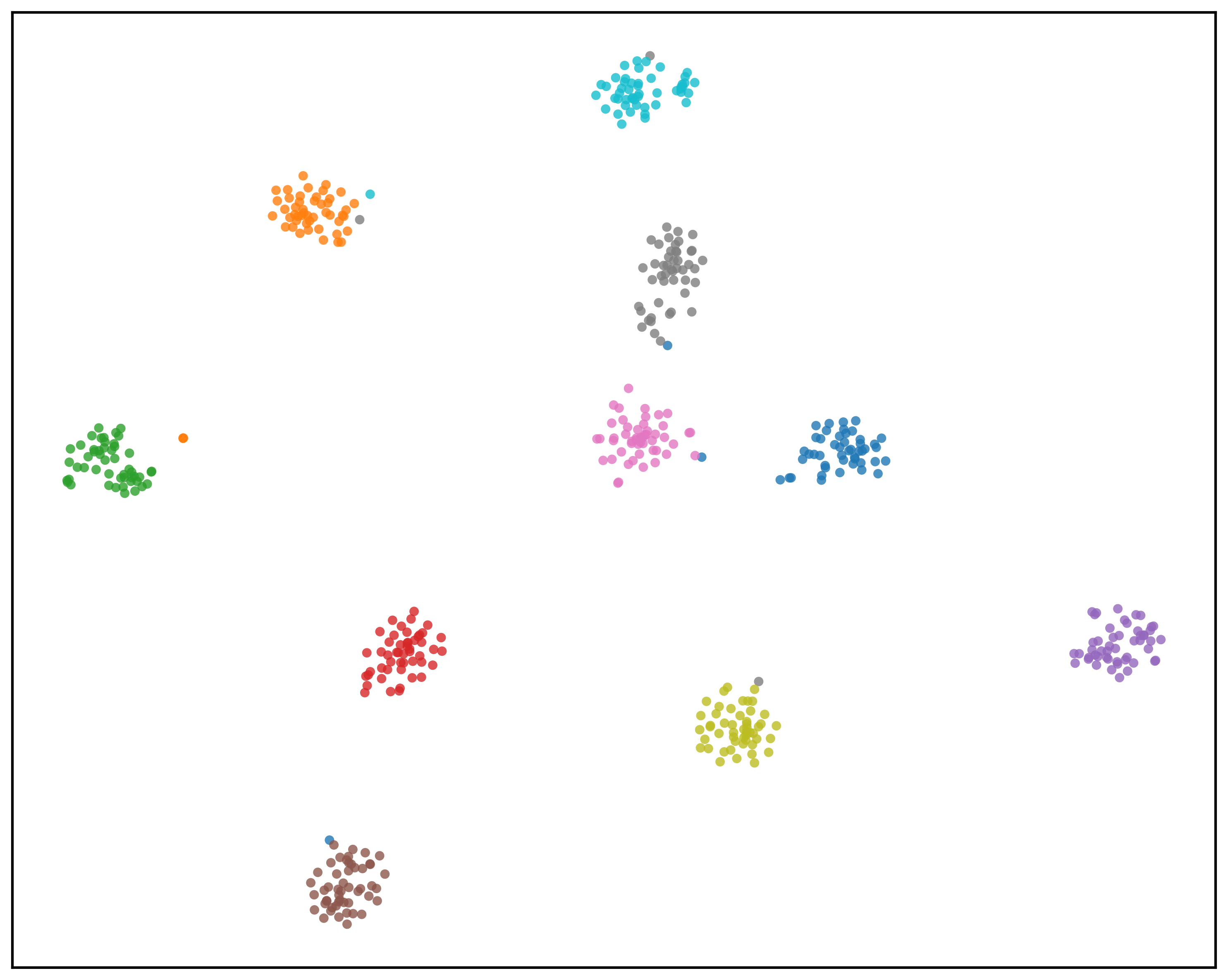}
        \caption{Raw Features (768D)}
        \label{fig:tsne_raw}
    \end{subfigure}
    \begin{subfigure}{0.49\linewidth}
        \centering
        \includegraphics[width=1.0\linewidth]{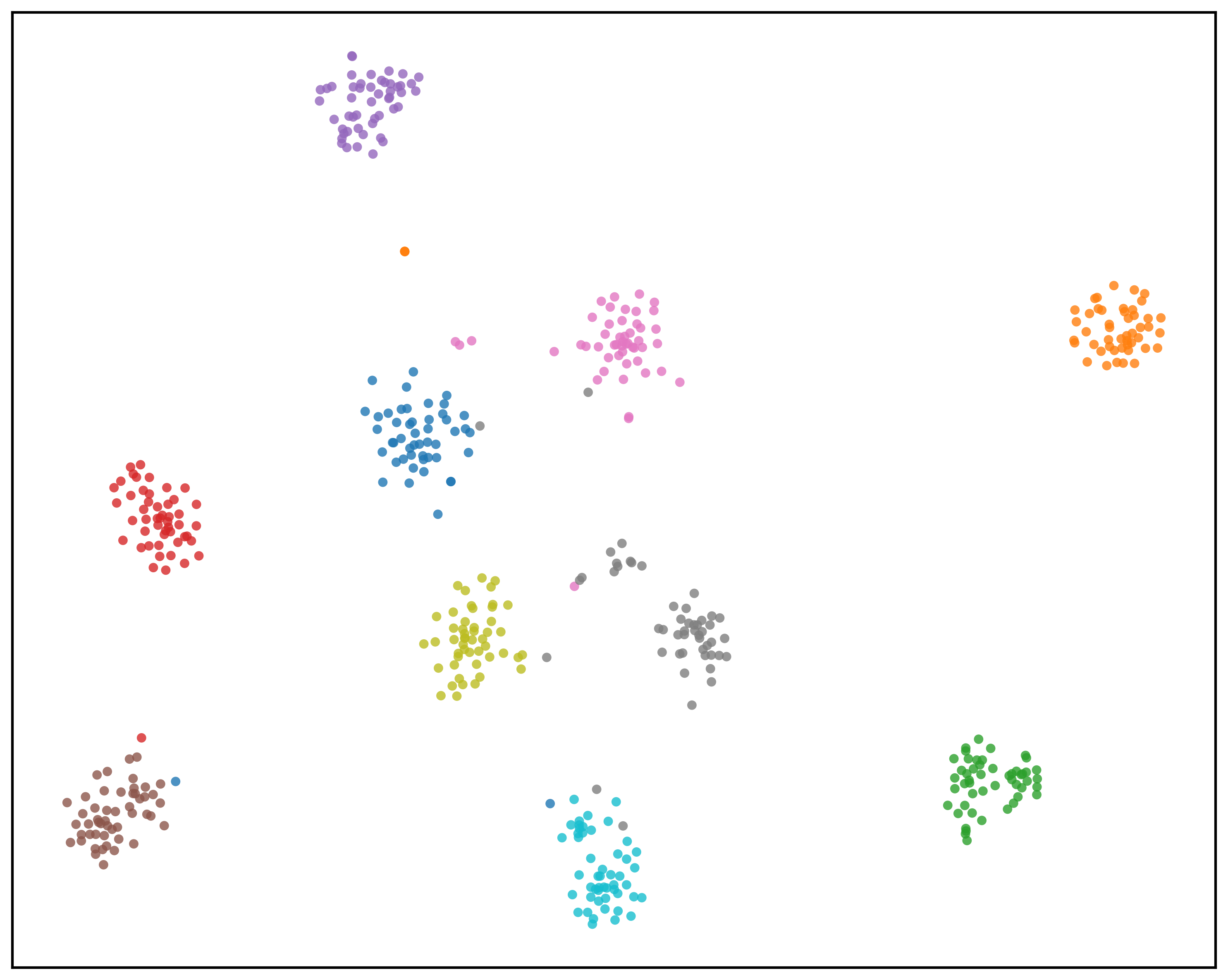}
        \caption{JL Projection (128D)}
        \label{fig:tsne_jl}
    \end{subfigure}
    \caption{$t$-SNE visualization of the same examples before and after a frozen JL projection. The same seed and the same $t$-SNE hyperparameters have been used across all visualizations.}
    \label{fig:tsne_raw_vs_jl}
\end{figure}

To provide a qualitative intuition, we visualize the feature space geometry using $t$-SNE in Figure~\ref{fig:tsne_raw_vs_jl}. Figure~\ref{fig:tsne_raw} displays the raw, high-dimensional representations ($d=768$) of the fully fine-tuned model, exhibiting clear class separation. Figure~\ref{fig:tsne_jl} shows the same samples projected into the random low-dimensional subspace ($k=128$) via the fixed JL matrix. 

Crucially, the cluster structure remains intact for the JL projection. The distinct separation between classes is preserved, and the local neighborhood relationships appear undisturbed. This supports our hypothesis that the essential geometry required for classification resides in a subspace that survives data-independent random contraction.

\section{Ablation}
To validate that the solution geometry is intrinsic to the data manifold rather than an artifact of specific projection techniques, we compare our oblivious JL approach against two distinct and data-dependent baselines:

\textbf{PCA projection.} We compute the principal components of the frozen training features and project the data onto the top-$k$ components. This represents the theoretically optimal linear projection for maximizing variance preservation.\\
\textbf{Learned projection.} Instead of a fixed random matrix $R$, we treat the projection matrix as a trainable parameter, which is initialized randomly to the same values as in the JL projection. This is updated end-to-end alongside the classifier head via backpropagation.

\textbf{Results.} Table \ref{tab:ablation} presents the comparative analysis across the architectures. We reproduce the JL results from Table \ref{tab:main_results} for direct comparison. As expected, PCA generally yields the highest accuracy because it is data-dependent, such that it aligns the projection basis with the axes of maximum information. However, the oblivious JL projection remains competitive to both ablations, the PCA projection, as well as the learned projection. This parity validates the 'Heavy-Tailed' spectral hypothesis \citet{martin2021implicit}: it implies that the discriminative signal is so structurally dominant over the noise bulk that precise eigen-alignment (via PCA) is redundant. The solution geometry is sufficiently robust to survive blind, random contraction.

\textbf{Vision (ResNet-50).} While PCA is the upper bound, JL projections remain on par. Notably, at $k=512$, JL achieves 81.20\% compared to 80.56\% for the learned matrix. Since the learned matrix was initialized from the exact same JL values, this performance drop suggests that optimizing the basis on the training set led to overfitting, whereas the frozen random features preserved the global geometry robustly.
\textbf{Transformers (ViT \& BERT).} In the Transformer domain, we observe a similar pattern. Notably, for ViT at $k=256$, JL outperforms both PCA and learned projections. 

\begin{figure}[t]
    \centering
    \begin{subfigure}{0.49\linewidth}
        \centering
        \includegraphics[width=1\linewidth]{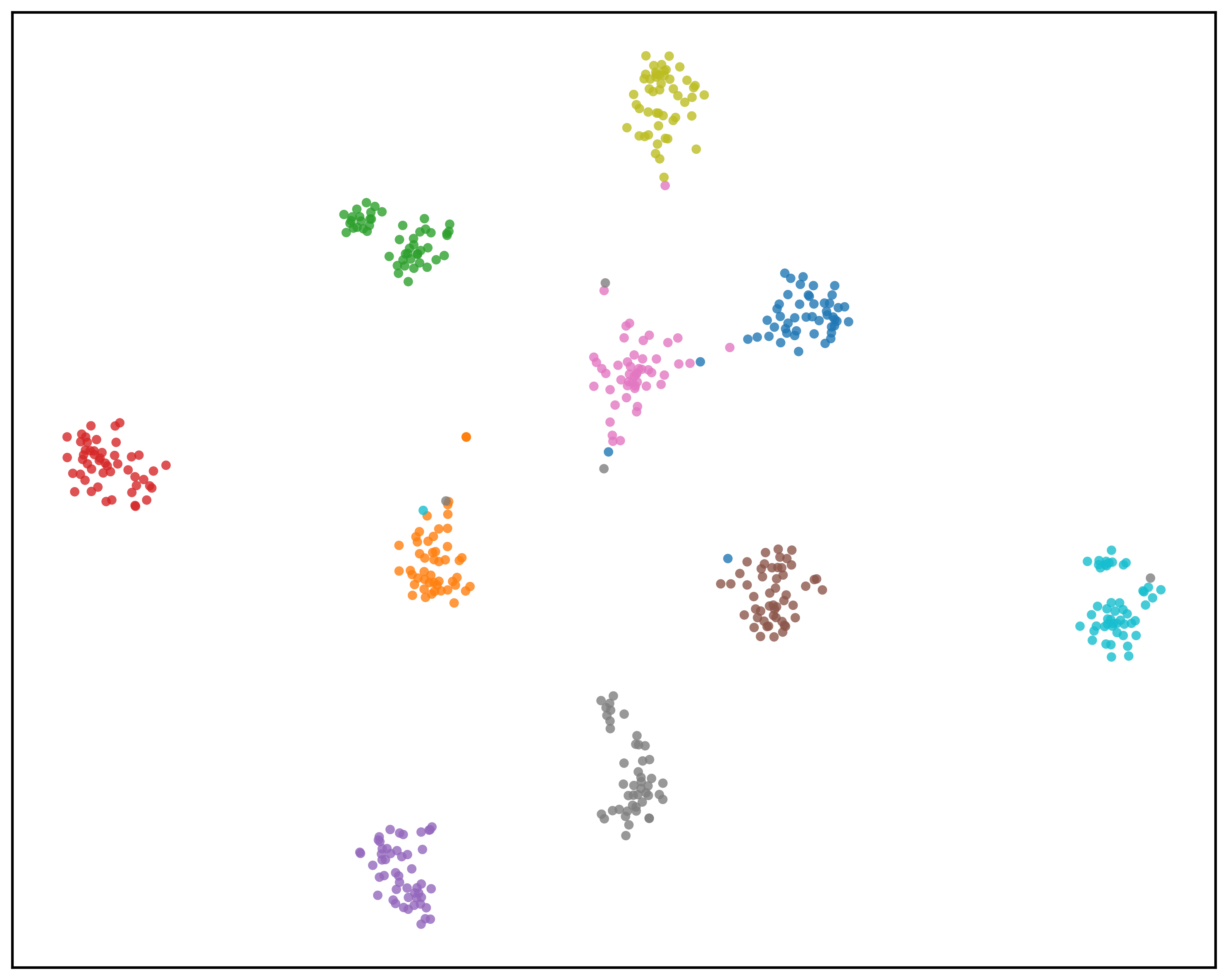}
        \caption{PCA Projection (128D)}
        \label{fig:tsne_pca}
    \end{subfigure}
    \begin{subfigure}{0.49\linewidth}
        \centering
        \includegraphics[width=1\linewidth]{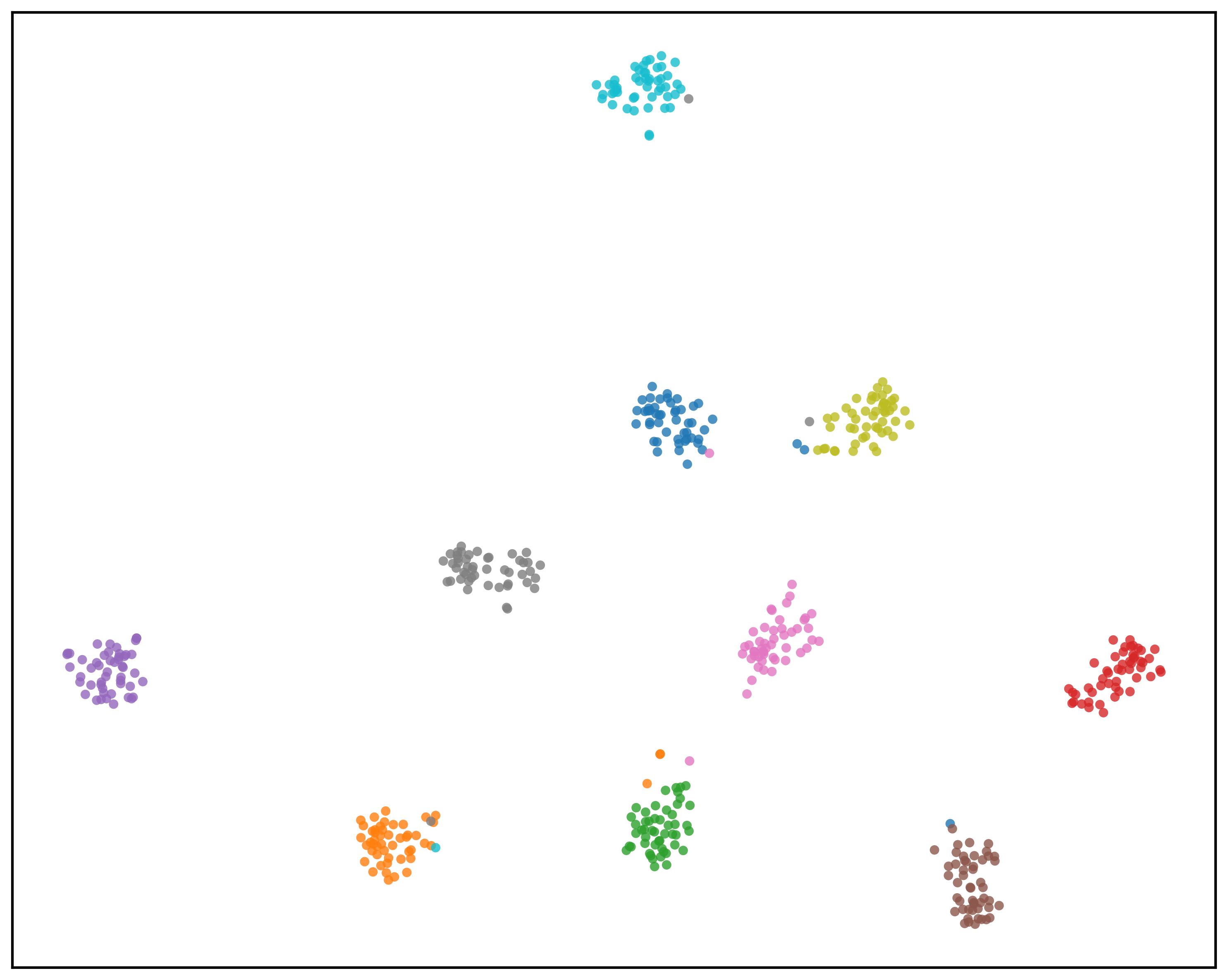}
        \caption{Learned Projection (128D)}
        \label{fig:tsne_learned}
    \end{subfigure}
    \caption{$t$-SNE visualization of the same examples for PCA projection and learned projection. The same seed and the same $t$-SNE hyperparameters have been used across all visualizations.}
    \label{fig:tsne_pca_vs_learned}
\end{figure}

We further compare the manifold structure of the subspaces in Figure~\ref{fig:tsne_pca_vs_learned}. While $t$-SNE introduces inherent non-linear distortions to the cluster shapes, the topological structure produced by the PCA projection (Figure~\ref{fig:tsne_pca}) and the learned projection (Figure~\ref{fig:tsne_learned}) is qualitatively comparable to our oblivious JL projection (Figure~\ref{fig:tsne_jl}).

\begin{table}[t]
\centering
\caption{\textbf{Ablation Study.} Comparison of JL projections (ours) against data-dependent (PCA) and optimization-dependent (Learned) projections. JL projections achieve parity with data-dependent methods, confirming the robustness of the solution geometry.}
\label{tab:ablation}
\setlength{\tabcolsep}{5pt}
\begin{tabular}{@{}l ccc@{}}
\toprule
Dim ($k$) & JL (Ours) & PCA & Learned \\ 
\midrule
\multicolumn{4}{@{}l}{\textit{ResNet-50 (CIFAR-100)}} \\ 
\midrule
1024 ($2\times$) & 81.32\% & \textbf{81.42}\% & 81.31\% \\
512 ($4\times$)  & 81.20\% & \textbf{81.30}\% & 80.56\% \\
256 ($8\times$)  & 80.89\% & \textbf{81.35}\% & 80.96\% \\
128 ($16\times$) & 80.19\% & \textbf{81.21}\% & 80.51\% \\ 
\midrule
\multicolumn{4}{@{}l}{\textit{ViT-B/16 (ImageNet-100) }} \\ 
\midrule
512 ($1.5\times$) & 93.52\% & \textbf{93.92\%} & 93.74\% \\
256 ($3\times$)   & \textbf{93.98}\% & 93.90\% & 93.78\% \\
128 ($6\times$)   & 93.90\% & \textbf{94.06\%} & 93.74\% \\
64 ($12\times$)   & 93.28\% & \textbf{93.98\%} & 93.42\% \\ 
\midrule
\multicolumn{4}{@{}l}{\textit{BERT-base (MNLI)}} \\ 
\midrule
512 ($1.5\times$) & 83.93\% & \textbf{83.96\%} & 83.83\% \\
256 ($3\times$)   & 83.79\% & 83.83\% & \textbf{83.88\%} \\
128 ($6\times$)   & 83.79\% & 83.70\% & \textbf{83.72\%} \\
64 ($12\times$)   & 83.64\% & 83.70\% & \textbf{83.98\%} \\ 
\bottomrule
\end{tabular}
\end{table}

\section{Discussion}
\label{sec:discussion}

\subsection{Implications: The Flat Solution Manifold}
Our findings provide constructive evidence for the Neural Flattening hypothesis. While prior work measured the intrinsic dimension via local gradients or manifold estimation \citep{ansuini2019intrinsic}, our use of a global, data-independent linear projection offers a stronger guarantee: the solution manifold is not only low-dimensional locally but is also globally well-behaved (flat and linearly separable).

The logic is as follows: If the class manifolds were highly curved or entangled in the high-dimensional space, a random linear projection would inevitably lead to significant class overlap and performance collapse (violating the margin requirements for JLL). The fact that this collapse does not occur implies that the backbone effectively linearizes the problem into a subspace $k \ll d$. The feature extractor acts as a flattening mechanism, rendering the complex input data into a geometry so simple that even a random basis is sufficient to distinguish the classes.

\subsection{Future Direction: Subspace-Native Distillation}
Our experiments have empirically confirmed the existence of a robust solution subspace that captures nearly the entire performance of the teacher despite being a fraction of the size. This finding challenges current knowledge distillation paradigms. Since we have shown that the relevant signal exists in the projected subspace $\tilde{h} \propto R h$, the practically relevant next step—which we leave for future work—is to exploit this for training and inference. For that goal, we propose the idea of Subspace-Native Distillation.
In this framework, the student is trained to directly construct the solution in the low-dimensional manifold defined by the fixed matrix $R$. The loss function shifts to:
\begin{equation}
    \mathcal{L}_{subspace} = || h_{student} - R h_{teacher} ||^2
\end{equation}
By targeting the intrinsic dimension directly, we hypothesize that the student can bypass the high-dimensional search problem entirely. The backbone of the student only needs enough capacity to map inputs to the $k$-dimensional target, potentially allowing for extreme parameter reduction while maintaining the teacher's accuracy. This paves the way for a generation of subspace-native models that are optimized for the solution, not the search.

\subsection{Limitations}

\textbf{Layer-wise Analysis.} While our findings provide evidence for the disconnect between optimization width and solution geometry, our analysis focuses exclusively on the final latent representation. As established by \citet{ansuini2019intrinsic}, the intrinsic dimension of deep networks is not uniform across layers; it typically peaks in intermediate layers before collapsing towards the output. It remains an open question whether our ``optimization scaffolding'' hypothesis implies that intermediate layers \textit{must} remain high-dimensional to facilitate gradient flow and feature disentanglement, or if the solution manifold is already constructible at earlier stages.


\paragraph{Task Generalization.}
Furthermore, our evaluation is centered on discriminative classification tasks where the number of classes is
small relative to the model width ($C < d$). In this regime, theoretical frameworks like Neural Collapse
predict low-rank separability. However, tasks with continuous output spaces (e.g., Variational Autoencoders, diffusion models)
or massive output vocabularies (where $C \gg d$, such as LLMs) may inherently require higher-dimensional manifolds
to preserve information that oblivious random projections might obliterate.

\subsection{Conclusion}
While high-dimensional width is crucial for solving the non-convex optimization problem during training, the final solution resides in a much simpler, low-rank subspace. This distinction provides the theoretical and empirical foundation for Subspace-Native Distillation. By shifting the distillation target from the noisy ambient space to this valid solution manifold, we can potentially train student models that are efficient by design rather than by approximation.

Ultimately, our framework validates a clear architectural philosophy for the future of efficient deep learning: \textbf{Train Big} to leverage the optimization benefits of high-dimensional scaffolding, but \textbf{Deploy Small} by explicitly extracting and targeting the intrinsic solution that remains.

\bibliography{paper}

@article{kaplan2020scaling,
  title={Scaling laws for neural language models},
  author={Kaplan, Jared and McCandlish, Sam and Henighan, Tom and Brown, Tom B and Chess, Benjamin and Child, Rewon and Gray, Scott and Radford, Alec and Wu, Jeffrey and Amodei, Dario},
  journal={arXiv preprint arXiv:2001.08361},
  year={2020}
}

@inproceedings{dosovitskiy2020vit,
  title={An Image is Worth 16x16 Words: Transformers for Image Recognition at Scale},
  author={Dosovitskiy, Alexey and Beyer, Lucas and Kolesnikov, Alexander and Weissenborn, Dirk and Zhai, Xiaohua and Unterthiner, Thomas and Dehghani, Mostafa and Minderer, Matthias and Heigold, Georg and Gelly, Sylvain and others},
  booktitle={International Conference on Learning Representations},
  year={2021}
}

@article{arriaga2006algorithmic,
  title={An algorithmic theory of learning: Robust concepts and random projection},
  author={Arriaga, Rosa I and Vempala, Santosh},
  journal={Machine learning},
  volume={63},
  number={2},
  pages={161--182},
  year={2006},
  publisher={Springer}
}

@inproceedings{li2018visualizing,
  title={Visualizing the loss landscape of neural nets},
  author={Li, Hao and Xu, Zheng and Taylor, Gavin and Studer, Christoph and Goldstein, Tom},
  booktitle={Advances in Neural Information Processing Systems},
  volume={31},
  year={2018}
}

@article{martin2021implicit,
  title={Implicit self-regularization in deep neural networks: Evidence from random matrix theory and implications for learning},
  author={Martin, Charles H and Mahoney, Michael W},
  journal={Journal of Machine Learning Research},
  volume={22},
  number={165},
  pages={1--73},
  year={2021}
}

@inproceedings{sedghi2018singular,
  title={The singular values of convolutional layers},
  author={Sedghi, Hanie and Gupta, Vineet and Long, Philip M},
  booktitle={International Conference on Learning Representations},
  year={2019}
}

@inproceedings{idelbayev2020low,
  title={Low-rank compression of neural networks: Learning the rank of each layer},
  author={Idelbayev, Yerlan and Carreira-Perpi{\~n}{\'a}n, Miguel A},
  booktitle={Proceedings of the IEEE/CVF Conference on Computer Vision and Pattern Recognition},
  pages={8049--8059},
  year={2020}
}

@inproceedings{denton2014exploiting,
  title={Exploiting linear structure within convolutional networks for efficient evaluation},
  author={Denton, Emily L and Zaremba, Wojciech and Bruna, Joan and LeCun, Yann and Fergus, Rob},
  booktitle={Advances in Neural Information Processing Systems},
  volume={27},
  year={2014}
}

@inproceedings{hu2021lora,
  title={LoRA: Low-Rank Adaptation of Large Language Models},
  author={Hu, Edward J and Shen, Yelong and Wallis, Phillip and Allen-Zhu, Zeyuan and Li, Yuanzhi and Wang, Shean and Wang, Lu and Chen, Weizhu},
  booktitle={International Conference on Learning Representations},
  year={2022}
}

@inproceedings{frankle2018lottery,
  title={The Lottery Ticket Hypothesis: Finding Sparse, Trainable Neural Networks},
  author={Frankle, Jonathan and Carbin, Michael},
  booktitle={International Conference on Learning Representations},
  year={2019}
}

@inproceedings{jacot2018neural,
  title={Neural tangent kernel: Convergence and generalization in neural networks},
  author={Jacot, Arthur and Gabriel, Franck and Hongler, Cl{\'e}ment},
  booktitle={Advances in Neural Information Processing Systems},
  volume={31},
  year={2018}
}

@inproceedings{lee2019wide,
  title={Wide neural networks of any depth evolve as linear models under gradient descent},
  author={Lee, Jaehoon and Xiao, Lechao and Schoenholz, Samuel and Bahri, Yasaman and Novak, Roman and Sohl-Dickstein, Jascha and Pennington, Jeffrey},
  booktitle={Advances in Neural Information Processing Systems},
  volume={32},
  year={2019}
}

@article{hinton2015distilling,
  title={Distilling the knowledge in a neural network},
  author={Hinton, Geoffrey and Vinyals, Oriol and Dean, Jeff},
  journal={arXiv preprint arXiv:1503.02531},
  year={2015}
}

@inproceedings{phuong2019towards,
  title={Towards understanding knowledge distillation},
  author={Phuong, Mary and Lampert, Christoph},
  booktitle={International Conference on Machine Learning},
  pages={5142--5151},
  year={2019},
  organization={PMLR}
}

@inproceedings{romero2014fitnets,
  title={FitNets: Hints for Thin Deep Nets},
  author={Romero, Adriana and Ballas, Nicolas and Kahou, Samira Ebrahimi and Chassang, Antoine and Gatta, Carlo and Bengio, Yoshua},
  booktitle={International Conference on Learning Representations},
  year={2015}
}

@inproceedings{cho2019efficacy,
  title={On the efficacy of knowledge distillation},
  author={Cho, Jang Hyun and Hariharan, Bharath},
  booktitle={Proceedings of the IEEE/CVF International Conference on Computer Vision},
  pages={4794--4802},
  year={2019}
}

@article{johnson1984extensions,
  title={Extensions of Lipschitz mappings into a Hilbert space},
  author={Johnson, William B and Lindenstrauss, Joram},
  journal={Contemporary mathematics},
  volume={26},
  number={189-206},
  pages={1},
  year={1984}
}

@inproceedings{devlin2018bert,
  title={BERT: Pre-training of Deep Bidirectional Transformers for Language Understanding},
  author={Devlin, Jacob and Chang, Ming-Wei and Lee, Kenton and Toutanova, Kristina},
  booktitle={Proceedings of the 2019 Conference of the North American Chapter of the Association for Computational Linguistics: Human Language Technologies},
  pages={4171--4186},
  year={2019}
}

@inproceedings{williams2018broad,
  title={A Broad-Coverage Challenge Corpus for Sentence Understanding through Inference},
  author={Williams, Adina and Nangia, Nikita and Bowman, Samuel},
  booktitle={Proceedings of the 2018 Conference of the North American Chapter of the Association for Computational Linguistics: Human Language Technologies},
  pages={1112--1122},
  year={2018}
}

@inproceedings{tian2020contrastive,
  title={Contrastive multiview coding},
  author={Tian, Yonglong and Krishnan, Dilip and Isola, Phillip},
  booktitle={Computer Vision--ECCV 2020: 16th European Conference, Glasgow, UK, August 23--28, 2020, Proceedings, Part XI 16},
  pages={776--794},
  year={2020},
  organization={Springer}
}

@inproceedings{he2016deep,
  title={Deep residual learning for image recognition},
  author={He, Kaiming and Zhang, Xiangyu and Ren, Shaoqing and Sun, Jian},
  booktitle={Proceedings of the IEEE Conference on Computer Vision and Pattern Recognition},
  pages={770--778},
  year={2016}
}

@techreport{krizhevsky2009learning,
  title={Learning multiple layers of features from tiny images},
  author={Krizhevsky, Alex and Hinton, Geoffrey and others},
  institution={University of Toronto},
  year={2009}
}

@article{donoho2006compressed,
  title={Compressed sensing},
  author={Donoho, David L},
  journal={IEEE Transactions on Information Theory},
  volume={52},
  number={4},
  pages={1289--1306},
  year={2006},
  publisher={IEEE}
}

@article{papyan2020prevalence,
  title={Prevalence of neural collapse during the terminal phase of deep learning training},
  author={Papyan, Vardan and Han, XY and Donoho, David L},
  journal={Proceedings of the National Academy of Sciences},
  volume={117},
  number={40},
  pages={24652--24663},
  year={2020}
}

@inproceedings{ansuini2019intrinsic,
  title={Intrinsic dimension of data representations in deep neural networks},
  author={Ansuini, Alessio and Laio, Alessandro and Macke, Jakob H and Zoccolan, Davide},
  booktitle={Advances in Neural Information Processing Systems},
  volume={32},
  year={2019}
}

@inproceedings{pope2021intrinsic,
  title={The Intrinsic Dimension of Images and Its Impact on Learning},
  author={Pope, Phil and Zhu, Chen and Abdelkader, Ahmed and Goldblum, Micah and Goldstein, Tom},
  booktitle={International Conference on Learning Representations},
  year={2021}
}

@inproceedings{li2018measuring,
  title={Measuring the Intrinsic Dimension of Objective Landscapes},
  author={Li, Chunyuan and Farkhoor, Heerad and Liu, Rosanne and Yosinski, Jason},
  booktitle={International Conference on Learning Representations},
  year={2018}
}

@inproceedings{rahimi2007random,
  title={Random features for large-scale kernel machines},
  author={Rahimi, Ali and Recht, Benjamin},
  booktitle={Advances in Neural Information Processing Systems},
  volume={20},
  year={2007}
}

@article{pearson1901lines,
  title={LIII. On lines and planes of closest fit to systems of points in space},
  author={Pearson, Karl},
  journal={The London, Edinburgh, and Dublin philosophical magazine and journal of science},
  volume={2},
  number={11},
  pages={559--572},
  year={1901},
  publisher={Taylor \& Francis}
}

@article{grattafiori2024llama,
  title={The llama 3 herd of models},
  author={Grattafiori, Aaron and Dubey, Abhimanyu and Jauhri, Abhinav and Pandey, Abhinav and Kadian, Abhishek and Al-Dahle, Ahmad and Letman, Aiesha and Mathur, Akhil and Schelten, Alan and Vaughan, Alex and others},
  journal={arXiv preprint arXiv:2407.21783},
  year={2024}
}

@article{hinton2006reducing,
  title={Reducing the dimensionality of data with neural networks},
  author={Hinton, Geoffrey E and Salakhutdinov, Ruslan R},
  journal={Science},
  volume={313},
  number={5786},
  pages={504--507},
  year={2006}
}

@article{devries2017improved,
  title={Improved regularization of convolutional neural networks with cutout},
  author={DeVries, Terrance and Taylor, Graham W},
  journal={arXiv preprint arXiv:1708.04552},
  year={2017}
}

@inproceedings{loshchilov2017decoupled,
  title={Decoupled Weight Decay Regularization},
  author={Loshchilov, Ilya and Hutter, Frank},
  booktitle={International Conference on Learning Representations},
  year={2019}
}
\bibliographystyle{icml2026}




\end{document}